\documentclass{article}
\usepackage{spconf,amsmath,amssymb,graphicx,hyperref}
\usepackage{algorithm}
\usepackage{algorithmic}
\usepackage{booktabs}
\usepackage{multirow}
\usepackage{xcolor}
\usepackage{stfloats}   
\def\bm{\boldsymbol{\theta}}

\def\bw{\mathbf{w}}
\def\bq{\mathbf{q}}

\title{BAYESIAN OPTIMIZATION WITH KERNEL ENSEMBLES and Disagreement-based Acquisition FOR SOURCE
LOCALIZATION AND ACOUSTIC INVERSION}

\name{Heng Zhang$^\dag$ \qquad Haotian Xiang$^\dag$ \qquad Florian Meyer$^\ddag$ \qquad Qin Lu$^\dag$
}
\address{$^\dag$School of Electrical and Computer Engineering,
University of Georgia, Athens, GA, USA\\
$^\ddag$Scripps Institution of Oceanography , University of California, San Diego, CA, USA
}

\begin{document}
\maketitle

\begin{abstract}
Joint source localization and geoacoustic inversion requires optimizing an
objective built from an expensive normal mode propagation model. Bayesian
optimization (BO) with a Gaussian process (GP) surrogate can obtain accurate
parameter estimates within a limited number of forward model evaluations, but
its performance depends on the choice of kernel family. With few observations
in a seven-dimensional search space, no single kernel can be expected to perform
consistently well across individual inversions. To reduce this dependence, we
use a weighted ensemble of GPs with different kernel families, allowing the
surrogate to adapt to the observed objective without committing to one kernel
in advance. The ensemble is combined with an optimum-conditioned acquisition
function that determines where the expensive objective should be evaluated
next. Experiments on simulated and measured SWellEx-96 data show that the
resulting method achieves the lowest mean final objective among the considered
BO strategies and reduces parameter estimation error on most coordinates.
Ablation results further show that the ensemble provides robustness to kernel choice, while the acquisition function accounts for most of the optimization gain.
\end{abstract}


\section{Introduction}
\label{sec:intro}

Estimating the position of an underwater source together with the geoacoustic
properties of the seabed is a fundamental inverse problem in ocean acoustics
\cite{chapman2021review}. Matched field inversion addresses this problem by
comparing measured array data with replica fields generated by an acoustic
propagation model \cite{baggeroer1993overview}. Conventional approaches search
the parameter space using grid-based, sampling-based, or population-based
optimization methods, including differential evolution and related stochastic
search strategies \cite{chapman2021review,jenkins2024geoacoustic}. These methods
can require thousands of forward model evaluations in a multidimensional
inversion problem, while each evaluation itself involves an acoustic propagation
calculation.

Bayesian optimization (BO) provides a more evaluation-efficient alternative by
fitting a Gaussian process (GP) surrogate to the observed objective and using an
acquisition function to determine where the objective should be evaluated next
\cite{shahriari2016taking,frazier2018tutorial}. BO has recently been applied to
source localization and geoacoustic inversion
\cite{jenkins2023bo,jenkins2024geoacoustic,jenkins2026turbo}, where accurate
parameter estimates can be obtained within approximately one hundred forward
model evaluations. Existing implementations, however, typically use a single GP
kernel family selected before the inversion. With only a small number of
observations in a seven-dimensional parameter space, there is no clear reason to
expect one kernel family to be consistently preferable.
Table~\ref{tab:kernelwin} illustrates this sensitivity. For BO strategies,
we repeat each run with three kernel families from an identical warm-up design,
so that the kernel is the only difference among the paired runs. Across 100
runs, no kernel family gives the lowest parameter error on more than half of the
runs. Moreover, the preferred family changes across strategies, runs, and data
sets. A fixed kernel selected in advance can therefore introduce a substantial
model-selection dependence into an individual inversion.

\begin{table}[t]
\centering
\caption{Percentage of runs on which each kernel family gives the lowest
normalized parameter error. Each run is repeated with all three families from
an identical warm-up design, so the three are directly paired; 100 runs per
row. No family is best on all the BO approaches.}
\label{tab:kernelwin}
\small
\setlength{\tabcolsep}{5pt}
\begin{tabular}{llccc}
\toprule
Strategy & Data & RBF & Mat\'ern-$3/2$ & Mat\'ern-$5/2$ \\
\midrule
\multirow{2}{*}{BO-LogEI}
  & Simulated & 42\% & 33\% & 25\% \\
  & Measured  & 49\% & 20\% & 31\% \\
\midrule
\multirow{2}{*}{Local BO $q{=}1$}
  & Simulated & 32\% & 29\% & 39\% \\
  & Measured  & 38\% & 22\% & 40\% \\
  \midrule
\multirow{2}{*}{Local BO $q{=}4$}
  & Simulated & 42\% & 18\% & 40\% \\
  & Measured  & 44\% & 21\% & 35\% \\
\bottomrule
\end{tabular}
\end{table}

To automate the kernel selection, we leverage a kernel ensemble to
maintain several plausible kernel families~\cite{egp,lu2023surrogate}. With each kernel inducing a GP model, this will yield a GP mixture based surrogate, whose weights are adapted to the collected evaluations, thus bypassing the need to commit to a single family
before the inversion begins. Then, a disagreement-based acquisition
function~\cite{scorebo} is adapted to select new evaluations. These two components play different roles
in the present problem: the ensemble reduces sensitivity to kernel selection,
while the acquisition improves the efficiency of the optimization itself. Evaluations on simulated and measured SWellEx-96 data corroborate the improved parameter estimation performance of the proposed method
over existing global and local BO baselines~\cite{jenkins2023bo,jenkins2024geoacoustic,jenkins2026turbo}.



\section{PROBLEM FORMULATION}
\label{sec:problem}
In geoacoustic inversion, a model vector
$\bm\in\Theta\subset\mathbb{R}^{D}$, containing the source-location and other parameters to be estimated, yields the {\it observed} complex pressure
field $\bq_{l,k}\in\mathbb{C}^{J}$ recorded by a vertical line array (VLA) with $J$ elements at frequency
$\omega_l\in \Omega:=\{\omega_1,\ldots,\omega_L\}$ and snapshot $k$  according to the following forward model\footnote{Note that BO-based approaches do not use the forward model~\eqref{eq:measurement} to estimate $\bm$ directly. Rather, \eqref{eq:measurement} is only used to yield the Bartlett-based objective~\eqref{eq:bartlett}.}\vspace{-0.3cm} 
\begin{equation}
\bq_{l,k}=\bw_l(\bm)S(\omega_l)+\mathbf{e}_{l,k}, \qquad k=1,\ldots,K.
\label{eq:measurement}
\end{equation}\vspace{-0.6cm}\\
where $\bw_l(\bm)\in\mathbb{C}^{J}$ is the replica field generated by the
KRAKEN normal mode model \cite{porter1991kraken}, $S(\omega_l)$ is an
unknown source term, and $\mathbf{e}_{l,k}\in\mathbb{C}^{J}$ accounts for the measurement noise and other unmodeled uncertainties and is independent across snapshots. 

Given observations $\{\{\bq_{l,k}\}_{l=1}^{L}\}_{k=1}^K$, we will rely on the Bartlett processor to measure the normalized spatial match between the measured and replica fields, following matched field
inversion \cite{gerstoft1998ocean,mecklenbrauker2000objective}.
Specifically, the single-frequency Bartlett-based objective is\vspace{-0.4cm}
\begin{equation}
\phi_l(\bm)=
\mathrm{tr}\,\widehat{\mathbf{R}}_l-
\frac{\bw_l^{\mathsf H}(\bm)\widehat{\mathbf{R}}_l\bw_l(\bm)}
{\bw_l^{\mathsf H}(\bm)\bw_l(\bm)}
\label{eq:bartlett}
\end{equation}\vspace{-0.5cm}\\
where $(\cdot)^{\mathsf H}$ denotes the Hermitian transpose and $\widehat{\mathbf R}_l$ is the normalized sample covairance matrix obtained from $K$ pressure snapshots as
$\widehat{\mathbf R}_l:=\frac{1}{K}\sum_{k=1}^{K}
\widehat{\bq}_{l,k}\widehat{\bq}_{l,k}^{\mathsf H}$ with $
\widehat{\bq}_{l,k}:=\frac{\bq_{l,k}}{\|\bq_{l,k}\|}$.
Here, the normalization
makes the objective invariant to the unknown source amplitude, and smaller
$\phi_l(\bm)$ indicates better agreement between the measured and replica
fields. 

Multiplying over different frequencies in $\Omega$ yields the following optimization problem\vspace{-0.4cm}
\begin{equation}
\widehat{\bm}=\arg\min_{\bm\in\Theta} \phi(\bm) \qquad \phi(\bm)=\prod_{l=1}^{L}\phi_l(\bm).
\label{eq:multifreq}
\end{equation}\vspace{-0.5cm}\\
This is a nontrivial task since it is hard to write the objective explicitly and each evaluation of the objective is costly due to the need to propagate the forward model. To address such a challenging problem with few forward-model evaluations, BO has been adapted in~\cite{jenkins2023bo,jenkins2024geoacoustic,jenkins2026turbo}. For the maximization convention used in BO, we define
$f(\bm)=-\phi(\bm)$. Specifically, BO builds a probabilistic surrogate $p(f(\bm)|\mathcal{D}_t)$ for $f(\bm)$  based on the available evaluations $\mathcal{D}_t:=\{(\bm_\tau,y_\tau)\}_{\tau=1}^{t}$ ($y_\tau$ is the output of $f(\bm_\tau)$) to judiciously select the next query point $\bm_{\tau+1}$ by the optimizing so-termed acquisition function (AF)~\cite{garnett2023bayesian}. In~\cite{jenkins2023bo,jenkins2024geoacoustic,jenkins2026turbo}, the surrogate model is given by the GP with a {\it preselected} kernel function~\cite{Rasmussen2006gaussian}, whose choice is nontrivial for the highly versatile geoacoustic setting.


\section{BO WITH KERNEL ENSEMBLES AND OPTIMIZER-CONDITIONED ACQUISITION}
\label{sec:method}
To avoid selecting the kernel function of the GP in advance, we will leverage a weighted
ensemble of GPs, each associated with a different kernel and assessed by a data-adaptive weight~\cite{egp}. Then, a disagreement-based AF
\cite{Kendo,scorebo} is used to select the next query point. The resulting approach is termed as ``KEDA" hereafter, which is abbreviated for ``{\bf K}ernel {\bf E}semble with {\bf D}isagreement-based {\bf A}cquisition."


\subsection{Kernel ensemble surrogate}
\label{ssec:ensemble}

Let $\mathcal{K}:=\{\kappa_1,\ldots,\kappa_R\}$
denote a dictionary of $R$ kernel function~\cite{egp, lu2023surrogate}. For each kernel
$\kappa_r$, $r\in\{1,\ldots,R\}$, we fit an independent GP to
$\mathcal{D}_t$, with kernel hyperparameters
$\boldsymbol{\lambda}_r$ estimated by maximizing its marginal likelihood as\vspace{-0.2cm}
\begin{equation}
\hat{\boldsymbol{\lambda}}_r = \arg\max_{{\boldsymbol{\lambda}}_r} \log p(\mathcal{D}_t\mid r;{\boldsymbol{\lambda}}_r)\;. \label{eq:hyper}
\end{equation}\vspace{-0.6cm}\\
The predictive distribution is represented by the GP mixture\vspace{-0.4cm}
\begin{equation}
p(f(\bm)\mid\mathcal{D}_t)
=
\sum_{r=1}^{R}
w_t^r\,
\mathcal{N}\!\left(
f(\bm);
\mu_{r,t}(\bm),
\sigma_{r,t}^2(\bm)
\right),
\label{eq:mixture}
\end{equation}\vspace{-0.4cm}\\
where $\mu_{r,t}$ and $\sigma_{r,t}^2$ are the posterior mean and variance
of the GP associated with $\kappa_r$~\cite{Rasmussen2006gaussian}. The kernel weights are determined by
their marginal likelihoods $p(\mathcal{D}_t\mid r;\widehat{\boldsymbol{\lambda}}_r)$ evaluated at the estimated hyperparameters $\widehat{\boldsymbol{\lambda}}_r$ per kernel $\kappa_r$~\cite{lu2023surrogate}. Here, we additionally add  a small weight to GPs with weights close to $0$ to promote exploration, yielding\vspace{-0.4cm}
\begin{equation}
w_t^r
=
(1-\epsilon)
\frac{
p(\mathcal{D}_t\mid r;\widehat{\boldsymbol{\lambda}}_r)
}{
\sum_{j=1}^{R}
p(\mathcal{D}_t\mid j;\widehat{\boldsymbol{\lambda}}_j)
}
+
\frac{\epsilon}{R},
\label{eq:weights}
\end{equation}\vspace{-0.4cm}\\
so that poorly weighted models are not removed completely during the finite
evaluation budget.

For later use in the AF, the mixture in \eqref{eq:mixture} is
approximated by a single GP via moment matching. The approximated GP is then $\check{p}(f(\bm)\mid\mathcal{D}_t)=\mathcal{N}(f(\bm);\bar{\mu}_t(\bm),\bar{\sigma}_t^2(\bm))$ with\vspace{-0.3cm}
\begin{align}
\bar{\mu}_t(\bm)
&=\!
\sum_{r=1}^{R}
w_t^r\mu_{r,t}(\bm),
\label{eq:mm-mean}
\\
\bar{\sigma}_t^2(\bm)
&=\!
\sum_{r=1}^{R}
w_t^r\sigma_{r,t}^2(\bm)
\!+\!
\sum_{r=1}^{R}\!
w_t^r
\left(
\mu_{r,t}(\bm)\!-\!\bar{\mu}_t(\bm)
\right)^2 .
\label{eq:mm-var}
\end{align}\vspace{-0.4cm}\\
The first term in \eqref{eq:mm-var} averages the uncertainty within the
individual GPs, while the second accounts for variation among their posterior
means.

\begin{figure*}[!t]
\centering
\includegraphics[trim=10bp 6bp 11bp 9bp,clip,width=0.99\textwidth]{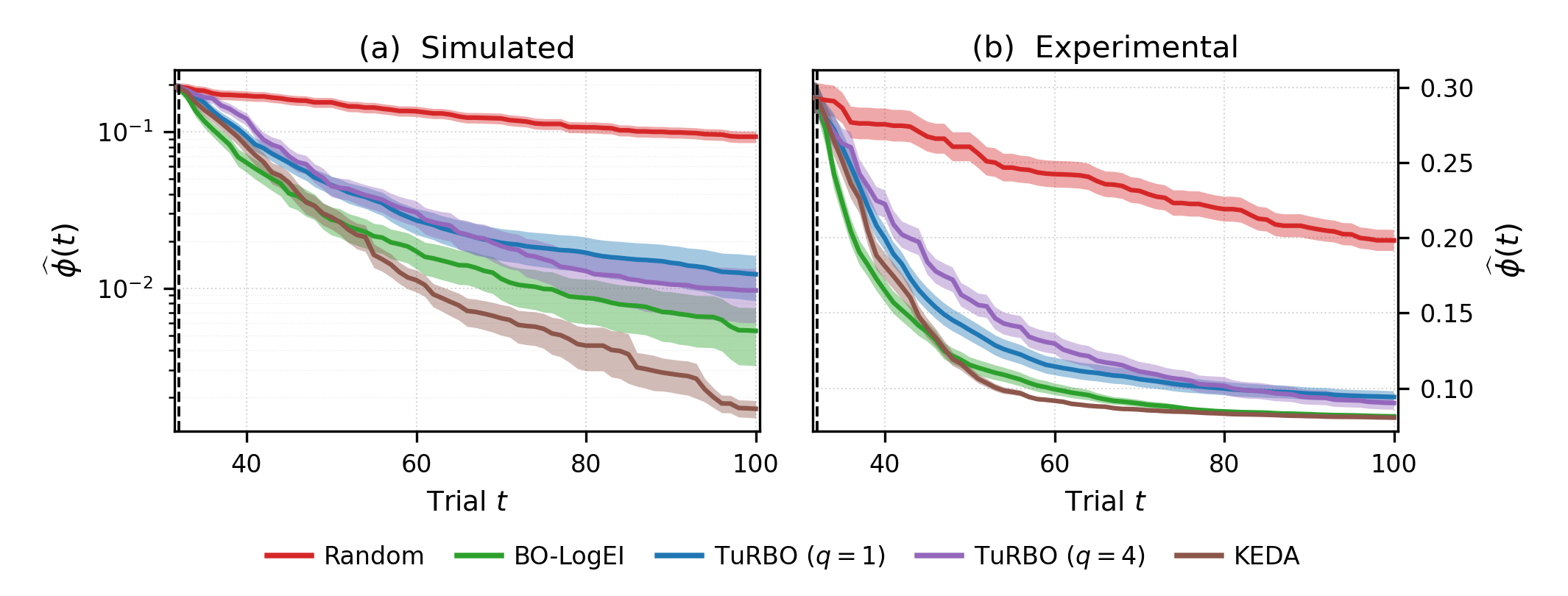}
\vspace{-10pt}
\caption{\small Lowest observed objective $\widehat{\phi}_t$ against trial index for (a) simulated and
(b) experimental data over 100 runs. Solid lines are the mean over runs, and
dash-dot lines of the same colour the 10th and 90th percentile at that trial.
The dots at the right edge show the distribution of the 100 final values.
Curves separate only after the $T_{\mathrm{init}}=32$ warm-up trials (dashed
vertical line), which are shared by all strategies at a given seed.}
\label{fig:convergence}
\end{figure*}

\subsection{Disagreement-based acquisition}
\label{ssec:acq}
Having available the GP ensemble-based surrogate, we are ready to design the AF. Here, we will adapt the disagreement-based rule~\cite{scorebo} to obtain the AF\vspace{-0.3cm}
\begin{equation}
\alpha_t(\bm) = \mathbb{E}_{r,{\bf \star}} \left[d(p(f(\bm)|{\cal D}_t), p(f(\bm)|r, {\bf \star},{\cal D}_t) ) \right]
\label{eq:AF}
\end{equation}\vspace{-0.5cm}\\
where $d(\cdot,\cdot)$ is a statistical distance (e.g, KL, Hellinger, Wasserstein) between two distributions, and ${\bf \star}:= (\bm^*,f^*)$ is the pair of the optimizer $\bm^*$ and the corresponding optimal function value $f^*$. A large value of $\alpha_t(\bm)$ therefore indicates a location whose
prediction is strongly affected by plausible hypotheses about the optimizer and the kernel function.

To evaluate the expectation in~\eqref{eq:AF}, we will adopt the posterior sampling. Specifically, for each GP/kernel $r$, we draw $S$ function posterior samples $\{\tilde{f}_{r,s}(\bm)\}_{s=1}^S$ using random
Fourier features \cite{rahimi2007random}. For sample $s$, its optimizer and
optimal value are\vspace{-0.3cm}
\begin{equation}
\tilde{\bm}^*_{r,s}
=
\arg\max_{\bm\in {\Theta}} \tilde{f}_{r,s}(\bm),
\qquad
\tilde{f}^*_{r,s}
=
\tilde{f}_{r,s}(\tilde{\bm}^*_{r,s}).
\label{eq:sample-opt}
\end{equation}\vspace{-0.5cm}\\
The pair $(\tilde{\bm}^*_{r,s},\tilde{f}^*_{r,s})$ represents one posterior hypothesis about
the optimum. Conditioning GP $r$ on this {\it phantom} observation gives the updated moments $\mu^{*_{s}}_{r,t}(\bm)$ and $\sigma^{2,*_{s}}_{r,t}(\bm)$.
Then, AF~\eqref{eq:AF} can be approximated as\vspace{-0.4cm}
\begin{equation}
\alpha_t(\bm)
\approx
\sum_{r=1}^{R}
\sum_{s=1}^{S}
\frac{w_t^r}{S}
\,d_{r,t}^s(\bm),
\label{eq:acq}
\end{equation}\vspace{-0.3cm}\\
where $d_{r,t}^s (\bm):=d(p(f(\bm)|{\cal D}_t), p(f(\bm)|r, \tilde{\bm}^*_{r,s},\tilde{f}^*_{r,s},{\cal D}_t) )$. To evaluate~\eqref{eq:acq} with tractability, we will approximate $p(f(\bm)|{\cal D}_t)$ by a single GP $\check{p}(f(\bm)|{\cal D}_t)$ with moments as in~\eqref{eq:mm-mean}-\eqref{eq:mm-var}. Specifically, for the squared Hellinger distance used in the experiments, we have the closed-form expression for $d_{r,t}^s (\bm)$ as\vspace{-0.4cm}
\begin{align}
d_{r,t}^{s}(\bm)
=
1
&-
\sqrt{
\frac{
2\bar{\sigma}_t(\bm)
\sigma^{*_{s}}_{r,t}(\bm)
}{
\bar{\sigma}_t^2(\bm)
+
\sigma^{2,*_{s}}_{r,t}(\bm)
}
}
\nonumber\\
&\times
\exp\!\left[
-\frac{
\left(
\bar{\mu}_t(\bm)
-
\mu^{*_{s}}_{r,t}(\bm)
\right)^2
}{
4\left(
\bar{\sigma}_t^2(\bm)
+
\sigma^{2,*_{s}}_{r,t}(\bm)
\right)
}
\right].
\label{eq:hellinger}
\end{align}\vspace{-0.4cm}

To obtain the next query point $\bm_{t+1}$, we optimize \eqref{eq:acq} over a Sobol candidate set
\cite{sobol1967}, followed by bound-constrained L-BFGS-B
\cite{zhu1997lbfgsb}.

\subsection{Complete procedure}
\label{ssec:algorithm}

Algorithm~\ref{alg:KEDA} summarizes the full inversion procedure. All GP
models rely on the same observations and differ only in their kernel family and
associated hyperparameters. Importantly, the ensemble requires no additional
acoustic forward-model evaluations: at each BO iteration only the selected
point $\bm_{t+1}$ is evaluated with KRAKEN to obtain the Bartlett objective~\eqref{eq:bartlett}.

\begin{algorithm}[t]
\caption{KEDA for matched field inversion}
\label{alg:KEDA}
\begin{algorithmic}[1]
\REQUIRE $\mathcal{K}$, $T_{\mathrm{total}}$, $T_{\mathrm{init}}$, $S$, $\epsilon$
\STATE Initialize $\mathcal{D}_{T_{\mathrm{init}}}$ with $T_{\mathrm{init}}$
Sobol points and $y_i=-\phi(\bm_i)$
\FOR{$t=T_{\mathrm{init}},\ldots,T_{\mathrm{total}}-1$}
    \STATE Fit each GP $r$ on ${\cal D}_t$ to obtain $\mu_{r,t}(\bm),
\sigma_{r,t}^2(\bm)$ with $\widehat{\boldsymbol{\lambda}}_r$ estimated by~\eqref{eq:hyper}; 
    \STATE Obtain weight $w_t^r$ according to \eqref{eq:weights}
    \STATE Form $\bar{\mu}_t(\bm),\bar{\sigma}_t^2(\bm)$ by
    \eqref{eq:mm-mean}--\eqref{eq:mm-var};
    \STATE Draw $\{\tilde{f}_{r,s}(\bm)\}$ and obtain
    $\{(\bm^*_{r,s},f^*_{r,s})\}$ via~\eqref{eq:sample-opt};
    \STATE Obtain $\mu^{*_{s}}_{r,t}(\bm)$ and $\sigma^{2,*_{s}}_{r,t}(\bm)$ using $\{(\bm^*_{r,s},f^*_{r,s})\}$;
    \STATE Evaluate $d_{r,t}^{s}(\bm)$ in~\eqref{eq:hellinger};
    \STATE Maximize~\eqref{eq:acq} to obtain $\bm_{t+1}$;
    \STATE Evaluate $\bm_{t+1}$ to get $y_{t+1}=-\phi(\bm_{t+1})$ based on~\eqref{eq:bartlett};
    \STATE Update
    $\mathcal{D}_{t+1}:=\mathcal{D}_{t}\cup \{(\bm_{t+1},y_{t+1}) \}$
\ENDFOR
\RETURN $\displaystyle
\arg\max_{(\bm_i,y_i)\in\mathcal{D}_{T_{\mathrm{total}}}} y_i$
\end{algorithmic}
\end{algorithm}

\begin{table}[t]
\centering
\caption{Search space $\Theta\subset\mathbb{R}^{D}$ in \eqref{eq:multifreq},
$D=7$}
\label{tab:searchspace}
\small
\setlength{\tabcolsep}{4pt}
\begin{tabular}{llc}
\toprule
Symbol & Parameter & Bounds \\
\midrule
$r_\mathrm{src}$ [km]  & Source range              & $[0.82,\ 1.32]$ \\
$z_\mathrm{src}$ [m]   & Source depth              & $[60,\ 80]$ \\
$\tau$ [$^\circ$]      & Array tilt                & $[-3,\ 3]$ \\
$h_w$ [m]              & Water depth               & $[212,\ 222]$ \\
$h_s$ [m]              & Sediment thickness        & $[15,\ 25]$ \\
$c_{s,t}$ [m/s]        & Sediment top speed        & $[1560,\ 1580]$ \\
$\delta c_s$ [m/s]     & Sediment speed increment  & $[10,\ 30]$ \\
\midrule
$c_{s,b}$ [m/s]        & $c_{s,t}+\delta c_s$ (reported) & $[1570,\ 1610]$ \\
\bottomrule
\end{tabular}
\end{table}

\begin{table*}[!t]
\centering
\caption{Mean final objective and MAE of the seven estimated parameters over 100 runs,
simulated / measured data. All baselines are our own reproduction under a
common warm-up design. Best value of each column and track in bold.}
\label{tab:mae}
\footnotesize
\setlength{\tabcolsep}{5pt}
\begin{tabular}{lcccccccc}
\toprule
Strategy &  ${\phi}^*$ & $r_\mathrm{src}$ [km] & $z_\mathrm{src}$ [m] & $\tau$ [$^\circ$] & $h_w$ [m] & $h_s$ [m] & $c_{s,t}$ [m/s] & $c_{s,b}$ [m/s] \\
\midrule
Best value & 0 / 0.076 & 1.07 / 1.072 & 71.11 / 71.53 & 2 / 2.068 & 217 / 219.6 & 23 / 20.07 & 1572.3 / 1567 & 1593 / 1590 \\
\midrule
Random & 0.0933 / 0.1983 & 0.028 / 0.035 & 2.04 / 1.96 & 0.392 / 0.399 & 2.63 / 3.00 & 3.15 / 2.53 & \textbf{5.24} / 5.62 & 6.47 / 5.96 \\
BO-LogEI & 0.0053 / 0.0813 & 0.018 / 0.012 & 0.81 / \textbf{0.27} & 0.195 / \textbf{0.041} & 1.98 / 1.42 & 1.81 / 1.73 & 6.64 / 7.55 & 7.03 / 6.24 \\
Local BO $q{=}1$ & 0.0123 / 0.0942 & 0.023 / 0.019 & 1.03 / 0.62 & 0.178 / 0.098 & 2.55 / 1.95 & 1.85 / 2.39 & 6.73 / 8.10 & 7.16 / 8.60 \\
Local BO $q{=}4$ & 0.0097 / 0.0902 & 0.023 / 0.015 & 1.05 / 0.62 & 0.168 / 0.101 & 2.30 / 1.22 & 1.53 / 1.39 & 5.65 / 6.25 & \textbf{5.92} / 5.75 \\
\textbf{KEDA} & \textbf{0.0017} / \textbf{0.0805} & \textbf{0.015} / \textbf{0.009} & \textbf{0.62} / \textbf{0.27} & \textbf{0.153} / 0.068 & \textbf{1.56} / \textbf{0.96} & \textbf{1.35} / \textbf{1.04} & 6.14 / \textbf{5.58} & 7.58 / \textbf{4.36} \\
\bottomrule
\end{tabular}
\end{table*}

\begin{figure*}[!t]
\centering
\includegraphics[trim=9bp 13bp 9bp 8bp,clip,width=\textwidth]{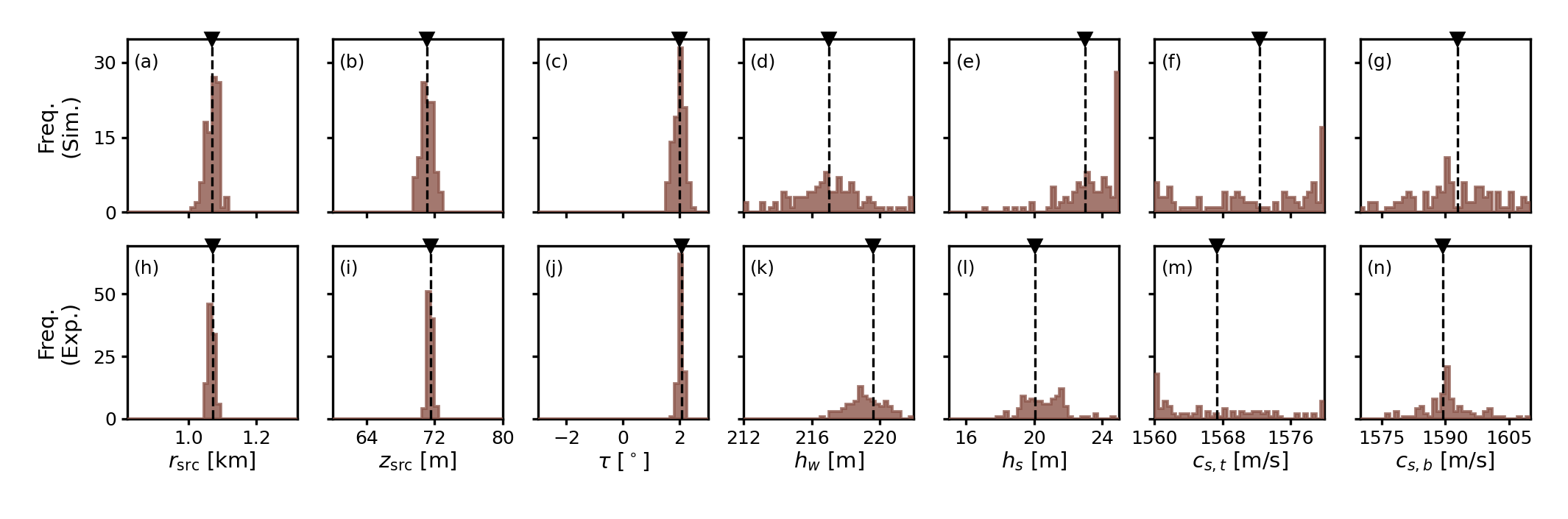}
\vspace{-10pt}
\caption{Distribution of the 100 final KEDA estimates for (a)--(g) simulated
and (h)--(n) measured data, 40 bins over the search bounds of each coordinate.
The dashed line and marker give the true value.}
\label{fig:params}
\end{figure*}

\section{Experimental Results}
\label{sec:results}

Localization and inversion are performed on simulated and real data from the
SWellEx-96 experiment \cite{swellex96}. Following the implementations in~\cite{jenkins2024geoacoustic,jenkins2026turbo}, the parameter vector $\bm$ to be estimated is 7-dimensional, whose physical meanings along with the feasible sets are summarized in Table~\ref{tab:searchspace}.
In addition, the VLA in \eqref{eq:measurement} has $J=21$ elements and $L=3$ tonals are
processed, and $\omega_l/2\pi\!\in\!\{148,\,235,\,388\}$\,Hz. On the real data,
$\widehat{\mathbf R}_l$ of \eqref{eq:bartlett} is formed from $K\!=\!8$ snapshots. For the simulated data, we set $K\!=\!1$ and consider the noise-free scenario, i.e., $\mathbf e_{l,k}=\mathbf 0$. Either way
$\mathrm{tr}\,\widehat{\mathbf R}_l=1$, so $\phi\in[0,1]$ in
\eqref{eq:multifreq}.  

Baselines include uniform random search, global BO with the log expected
improvement acquisition \cite{ament2023logei}, and the local BO strategy of
\cite{jenkins2026turbo} with $q=1$ and $q=4$, all with a Mat\'ern-5/2 kernel. Each strategy is given $T_{\mathrm{total}}=100$ evaluations with
$T_{\mathrm{init}}=32$ warm-up trials and is repeated over 100 random seeds. The initial query points
are drawn from a seeded Sobol sequence, so at a given seed every strategy starts
from the same design and differs only in how the remaining evaluations are
placed. 
For KEDA, the kernel dictionary consists of $R=3$ kernels, namely,  RBF, {Mat\'ern-}3/2, and {Mat\'ern-}5/2.
We set $S=8$ in ~\eqref{eq:acq} and $\epsilon=0.05$ in~\eqref{eq:weights}. Each $\tilde f_{r,s}$ of \eqref{eq:sample-opt} is realized with
$1024$ random Fourier features \cite{rahimi2007random} and its argmax found
from $128$ raw samples with $10$ restarts; $\alpha_t(\bm)$ \eqref{eq:acq}  is
maximized over $1024$ Sobol candidates \cite{sobol1967} with $40$ restarts.
{All the BO-based methods are implemented using BoTorch~\cite{balandat2020botorch}. 
}

{We evaluate optimization performance using the lowest observed objective
after trial $t$ as\vspace{-0.2cm}
\begin{equation}
\widehat{\phi}_t:=\min_{1\leq \tau \leq t}\phi(\bm_\tau).
\end{equation}\vspace{-0.4cm}\\
As shown in Fig.~\ref{fig:convergence}, KEDA achieves the lowest mean $\widehat{\phi}_t$ on both
data sets, with a pronounced improvement on simulated data and a smaller
advantage on real data.

Further, estimation performance has been evaluated via the mean absolute error (MAE). Specifically, the MAE for parameter $d$ over $N_{\mathrm{run}}=100$ runs is defined as\vspace{-0.3cm}
\begin{equation}
\mathrm{MAE}(\theta_d)
:=
\frac{1}{N_{\mathrm{run}}}
\sum_{i=1}^{N_{\mathrm{run}}}
\left|
\widehat{\theta}^{(i)}_d-\theta_{d}^{\mathrm{GT}}
\right|
\label{eq:mae}
\end{equation}\vspace{-0.5cm}\\
where $\widehat{\theta}^{(i)}_d$ is the $d$th component of the final
best estimate in run $i$, and $\theta_{d}^{\mathrm{GT}}$ is its ground truth value. Note that there is no ground truth of $\bm$ for real data, and we set $\theta_{d}^{\mathrm{GT}}$ to be the solution given by the differential evolution approach per
\cite{jenkins2024geoacoustic}. 
{The ``Best value'' row of Table~\ref{tab:mae} lists $\theta_{d}^{\mathrm{GT}}$ for both
scenarios, along with the optimal objective ${\phi}^*$. 
}
As shown in Table~\ref{tab:mae},  KEDA achieves the lowest or tied-lowest error
on five of the seven parameters for simulated data and six of the seven
parameters for real data. Fig.~\ref{fig:params} further shows that the
source range, source depth, and array tilt are tightly concentrated around
their reference values, while the environmental parameters exhibit broader
residual variation, particularly the sediment sound speeds.

\begin{figure*}[!t]
\centering
\includegraphics[trim=11bp 5bp 11bp 9bp,clip,width=\textwidth]{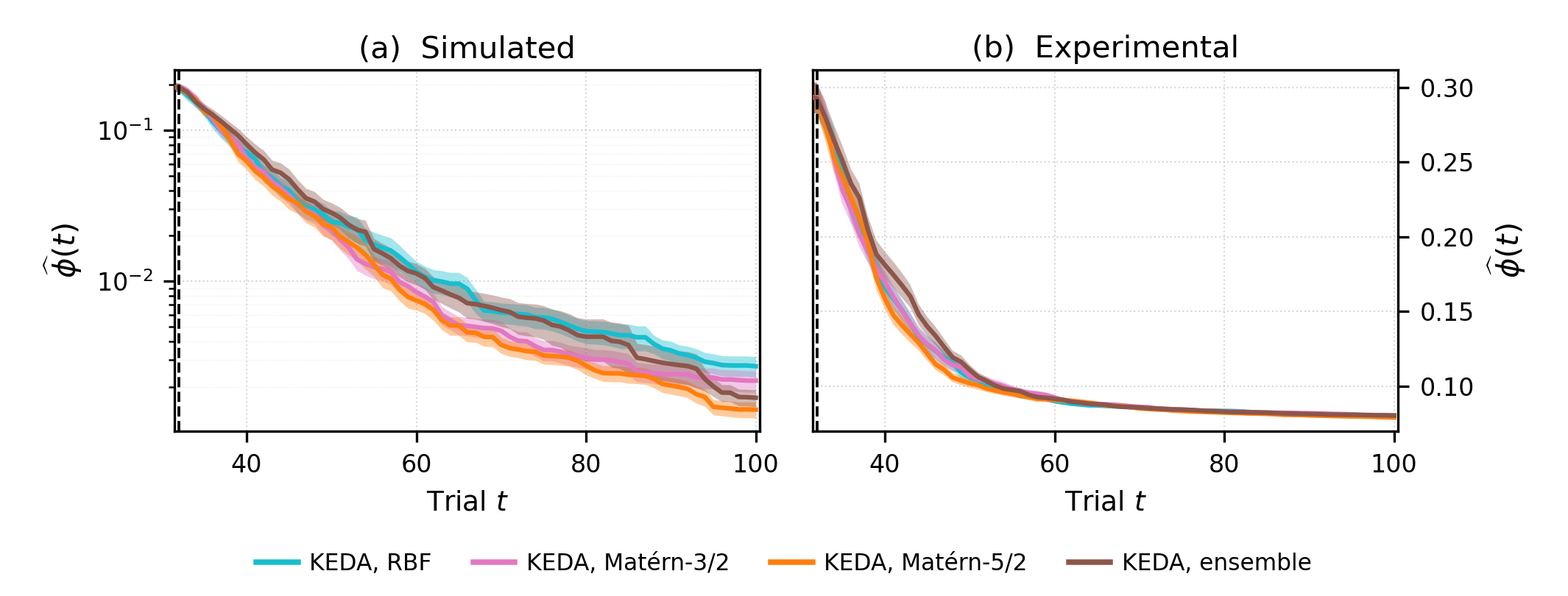}
\vspace{-6pt}
\caption{KEDA with each single kernel of $\mathcal{K}$ and with the full
ensemble, on (a) simulated and (b) experimental data, over 100 seeds.}
\label{fig:abl-KEDA}
\end{figure*}

\begin{figure*}[!t]
\centering
\includegraphics[trim=11bp 5bp 11bp 9bp,clip,width=\textwidth]{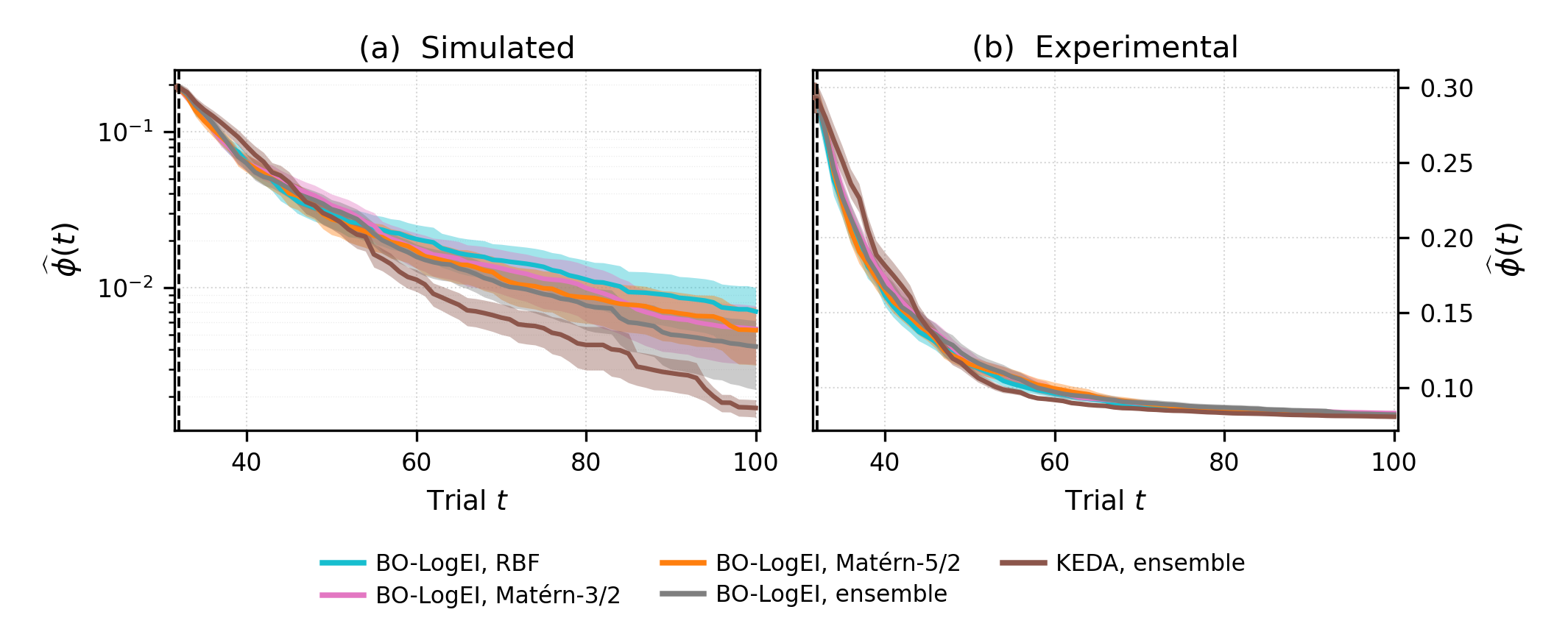}
\vspace{-6pt}
\caption{The same four surrogates under log expected improvement instead of the
disagreement acquisition.}
\label{fig:abl-logei}
\end{figure*}

\begin{figure*}[!t]
\centering
\includegraphics[trim=11bp 5bp 11bp 9bp,clip,width=\textwidth]{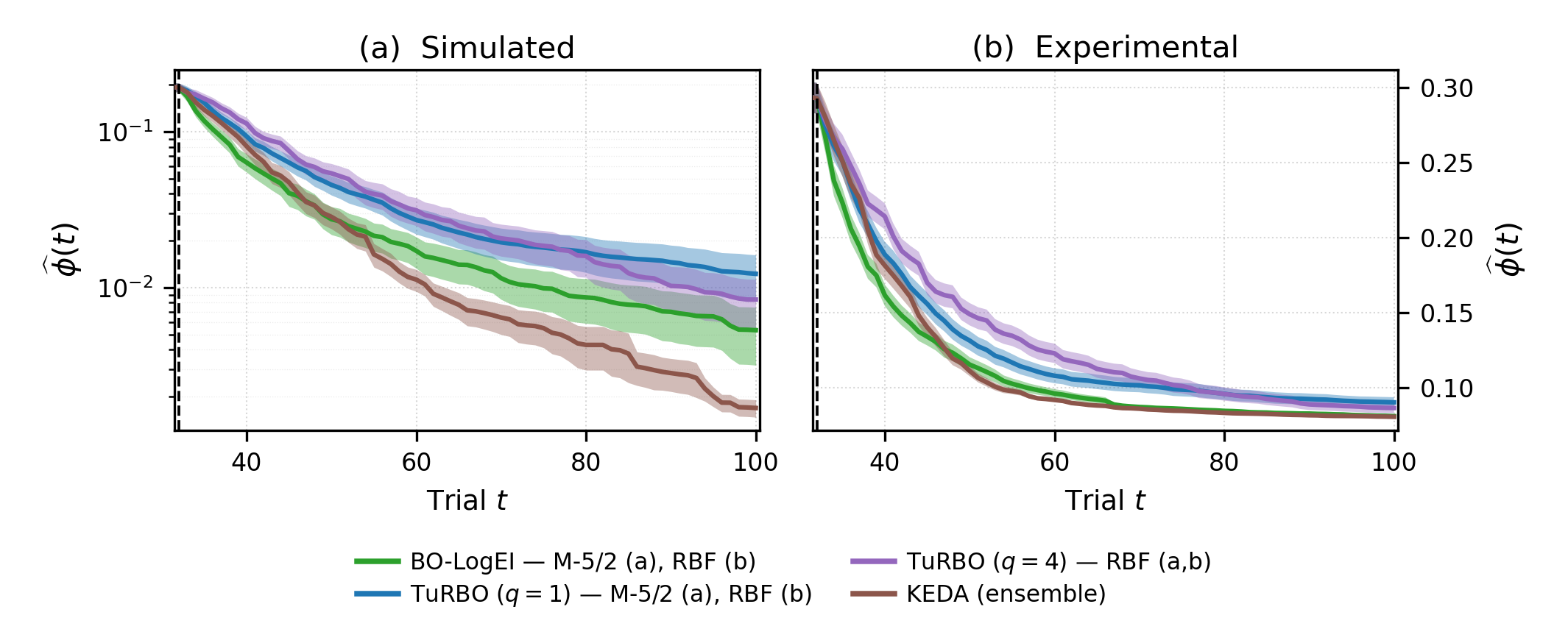}
\vspace{-6pt}
\caption{KEDA against each baseline at the kernel minimising that baseline's
own mean final objective; the legend names the kernel used in each panel.}
\label{fig:abl-best}
\end{figure*}
\begin{figure*}[!t]
\centering
\includegraphics[trim=11bp 2bp 8bp 9bp,clip,width=\textwidth]{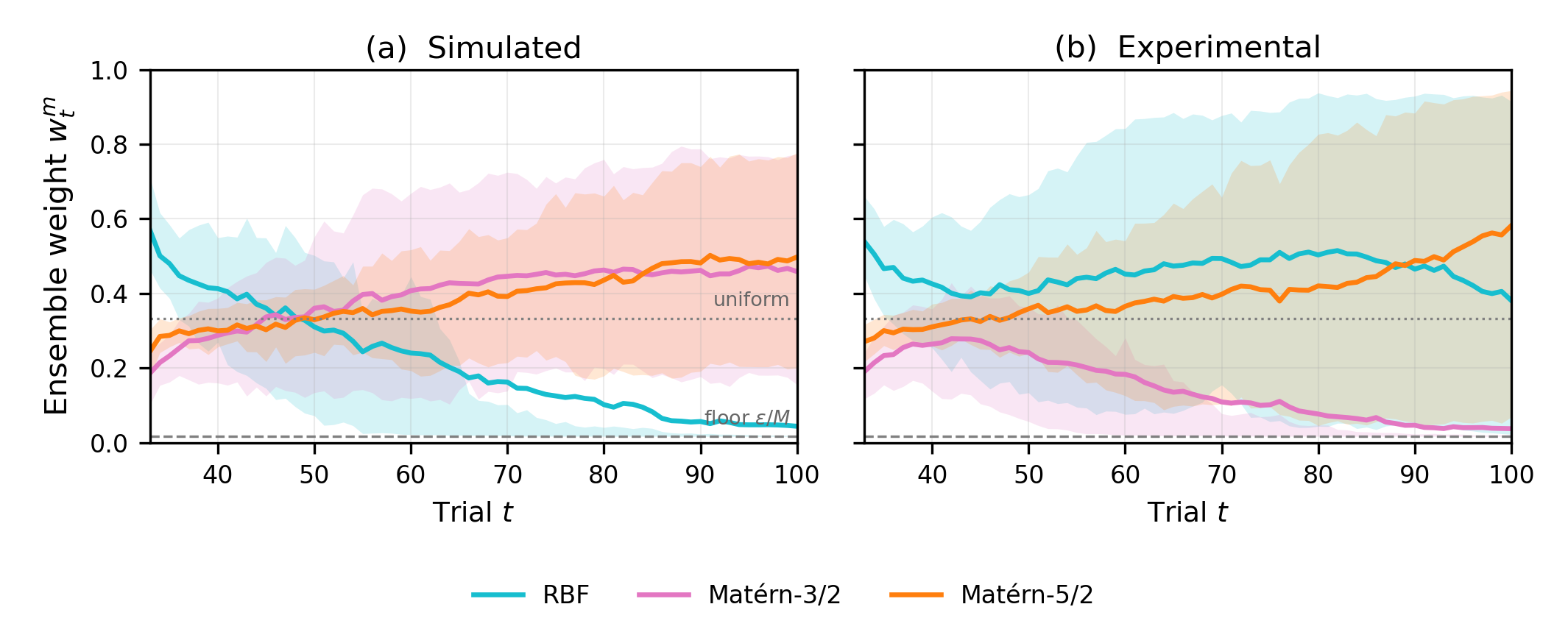}
\vspace{-6pt}
\caption{Kernel weights $w_t^r$ of \eqref{eq:weights} against trial $t$ on (a)
simulated and (b) measured data, over 100 seeds. Solid lines give the mean over
seeds and the shaded bands the interquartile range. The dotted line marks the
uniform weight $1/R$ and the dashed line the floor $\epsilon/R$. Weights exist
only after the warm-up, so the curves begin at $t=T_{\mathrm{init}}+1$.}
\label{fig:weights}
\end{figure*}

\begin{figure*}[!t]
\centering
\includegraphics[trim=6bp 12bp 8bp 9bp,clip,width=\textwidth]{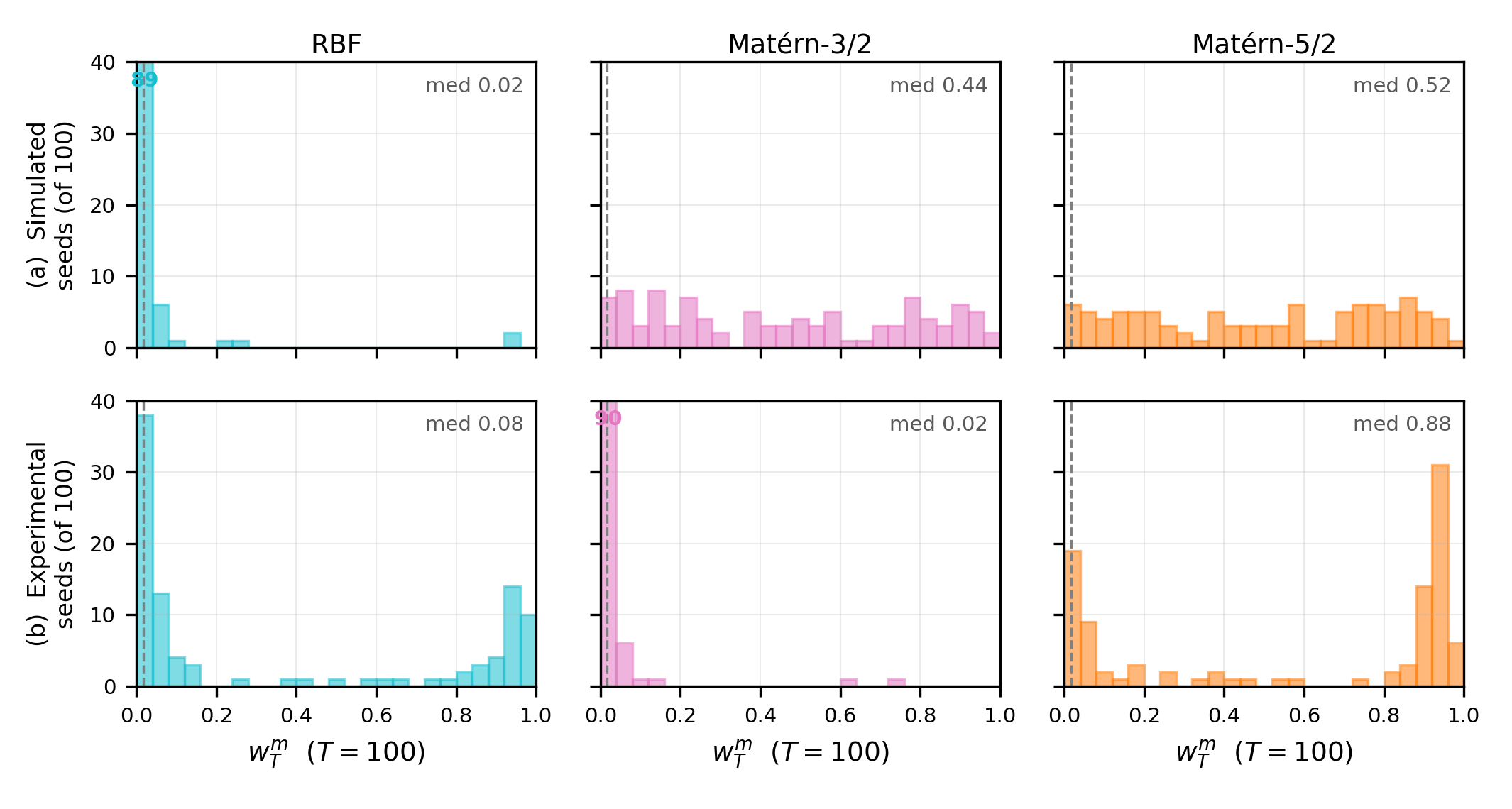}
\vspace{-6pt}
\caption{Distribution over the 100 seeds of the final weight $w_T^r$ of each
kernel in $\mathcal{K}$, on (a) simulated and (b) measured data. The dashed line
marks the floor $\epsilon/R$; bars exceeding the vertical range are labelled
with their count, and the median of each panel is given at its top right.}
\label{fig:weights-hist}
\end{figure*}

\subsection{Ablation Experiments}
\label{app:ablation}

Two ablations separate the ensemble from the acquisition. The first runs KEDA
with a single kernel, once for each family in $\mathcal{K}$; the second applies
log expected improvement to the moment matched posterior of the full ensemble.
Every baseline is also run with each family. 
The ablations distinguish the roles of the two KEDA components. The best single-kernel KEDA variant performs similarly to the full ensemble, showing
that the ensemble primarily removes the need to select a kernel family in
advance . In contrast,
replacing the optimizer-conditioned acquisition with LogEI
degrades performance, indicating that the acquisition accounts for most of
the improvement. KEDA also remains ahead of the baselines when each baseline
is given its retrospectively best fixed kernel, showing that the improvement
is not an artifact of an unfavorable kernel choice.


\noindent {\bf Removing the ensemble.} Fig.~\ref{fig:abl-KEDA} runs KEDA with a single kernel, once for each family
in $\mathcal{K}$, against the full ensemble. The four arms converge to the same
level on both data sets, and the spread among them is far smaller than the gap
separating any of them from the baselines in Fig.~\ref{fig:abl-best}. The
ensemble is therefore not what produces the result. What it does provide is
independence from the choice of family: it lands with the single-kernel arms
without that choice having to be made, at the cost of converging somewhat more
slowly than the best of them.

\noindent {\bf Giving the ensemble to the baseline.}
Fig.~\ref{fig:abl-logei} repeats the same comparison with logEI in place of the disagreement acquisition. The ensemble leaves the
baseline in the same band as its own single-kernel arms, and the whole family
stays above every arm of Fig.~\ref{fig:abl-KEDA} on both data sets. Read
together, the two figures place the improvement in the acquisition and not in
the surrogate: it survives every change of kernel and disappears as soon as the
acquisition is replaced.

\noindent{\bf Baselines at their own best kernel.}
Fig.~\ref{fig:abl-best} shows each baseline with the kernel that minimises its
own mean final objective on that data set. That kernel is not the same on the
two data sets, and not the same as the one preferred by KEDA, which is itself
evidence that the choice cannot be settled once and carried across problems.
KEDA stays ahead of every baseline even under this favourable choice, so the
margin reported in Section~\ref{sec:results} is not an artefact of holding the
baselines to a kernel that happens to suit them poorly.

\noindent{\bf Evolution of the Kernel Weights.}
Fig.~\ref{fig:weights} tracks the weights \eqref{eq:weights} over the run. On
both data sets the warm-up leaves most of the weight on the RBF, and on both
that ordering is overturned once evaluations accumulate, but not in the same
direction: simulated data drives the RBF to the floor and settles on the two
Mat\'ern families, whereas measured data discards Mat\'ern-$3/2$ and keeps the
RBF alongside Mat\'ern-$5/2$. Fig.~\ref{fig:weights-hist} shows the final
weights of the individual runs behind those means. On simulated data the two
surviving families are mixed over the whole range, so the ensemble is a genuine
average. On measured data the weights are bimodal: each run concentrates on
either the RBF or Mat\'ern-$5/2$, and the runs do not agree on which.

The family favoured by the evidence therefore depends on the data set, differs
from the one the warm-up design suggests, and on measured data is not even
settled within a single data set. Fixing a kernel before the run is a guess made
before the evidence that would decide it exists, and the ablations of
Appendix~\ref{app:ablation} show the baselines are held to exactly such a guess.
KEDA never makes it. The dictionary $\mathcal{K}$ is the same on both data
sets, nothing is tuned per data set, and the weights follow the evidence as it
arrives, which is how the results of Section~\ref{sec:results} are obtained.

\section{CONCLUSION}
\label{sec:conclusion}\vspace{-0.2cm}

We applied BO with weighted GP ensembles to joint
source localization and geoacoustic inversion. The ensemble reduces dependence
on selecting a single kernel family in advance, while the disagreement-based
acquisition provides most of the optimization gain. On simulated and real
SWellEx-96 data, KEDA achieves the lowest mean final objective among the
considered BO strategies and the lowest or tied-lowest parameter error on most
coordinates. Ablations further show that the ensemble performs comparably to
strong single-kernel variants without requiring that kernel to be chosen in
advance, while the advantage over the baselines persists when they are given
their retrospectively best fixed kernels. Extending the method to batch
evaluation and exploiting the product structure of the multi-frequency objective
in \eqref{eq:multifreq} are left for future work.

\clearpage

\bibliographystyle{IEEEbib}
\bibliography{refs}


\end{document}